%% file: main.tex
\documentclass{article}

\PassOptionsToPackage{numbers,compress}{natbib}
\usepackage[nonatbib, final]{neurips_2022}
\usepackage{listings}

\usepackage[utf8]{inputenc} 
\usepackage[T1]{fontenc}    
\usepackage{url}            
\usepackage{booktabs}       
\usepackage{amsfonts}       
\usepackage{nicefrac}       

\usepackage{listings}
\usepackage{microtype}      
\usepackage{xcolor}         
\usepackage{algorithm}
\usepackage[noend]{algpseudocode}
\input{math_commands.tex}
\usepackage{graphicx}
\usepackage[font=small]{caption}
\usepackage{wrapfig}
\usepackage{booktabs}
\usepackage{multirow}
\usepackage{enumitem}

\definecolor{citecolor}{HTML}{0071bc}
\usepackage{hyperref}
\hypersetup{pagebackref=true,breaklinks=true,colorlinks, citecolor=citecolor,linkcolor=black,bookmarks=false}
\usepackage[numbers]{natbib}
\usepackage{multirow}
\usepackage{nicefrac}
\newcommand{\xxnote}[3]{}
\renewcommand{\xxnote}[3]{\color{#2}{#1: #3}}

\newcommand{\method}{BeT}

\title{Behavior Transformers:\\Cloning $k$ modes with one stone}

\author{
Nur Muhammad (Mahi) Shafiullah
\thanks{Corresponding author, email: \texttt{mahi@cs.nyu.edu}}
\And Zichen Jeff Cui
\AND Ariuntuya Altanzaya
\And Lerrel Pinto
\AND New York University
}

\begin{document}

\maketitle
\input{abstract}
\input{introduction}

\input{method}
\input{results}
\input{related_work}

\input{conclusions}


\begin{ack}
We thank Ben Evans, David Brandfonbrener, Abitha Thankaraj, Jyo Pari, and Anthony Chen for valuable feedback and discussions.
This work was supported by grants from Honda, and ONR award numbers N00014-21-1-2404 and N00014-21-1-2758.
\end{ack}

\newpage
\bibliographystyle{abbrvnat}
\bibliography{references}
\newpage
\section*{Checklist}

The checklist follows the references.  Please
read the checklist guidelines carefully for information on how to answer these
questions.  For each question, change the default \answerTODO{} to \answerYes{},
\answerNo{}, or \answerNA{}.  You are strongly encouraged to include a {\bf
justification to your answer}, either by referencing the appropriate section of
your paper or providing a brief inline description.  For example:
\begin{itemize}
  \item Did you include the license to the code and datasets? \answerYes{}
  \item Did you include the license to the code and datasets? \answerNo{The code and the data are proprietary.}
  \item Did you include the license to the code and datasets? \answerNA{}
\end{itemize}
Please do not modify the questions and only use the provided macros for your
answers.  Note that the Checklist section does not count towards the page
limit.  In your paper, please delete this instructions block and only keep the
Checklist section heading above along with the questions/answers below.

\begin{enumerate}

\item For all authors...
\begin{enumerate}
  \item Do the main claims made in the abstract and introduction accurately reflect the paper's contributions and scope?
    \answerYes{}
  \item Did you describe the limitations of your work?
    \answerYes{See Sec.~\ref{sec:limit}.}
  \item Did you discuss any potential negative societal impacts of your work?
    \answerNA{}
  \item Have you read the ethics review guidelines and ensured that your paper conforms to them?
    \answerYes{}
\end{enumerate}

\item If you are including theoretical results...
\begin{enumerate}
  \item Did you state the full set of assumptions of all theoretical results?
    \answerNA{}
        \item Did you include complete proofs of all theoretical results?
    \answerNA{}
\end{enumerate}

\item If you ran experiments...
\begin{enumerate}
  \item Did you include the code, data, and instructions needed to reproduce the main experimental results (either in the supplemental material or as a URL)?
    \answerYes{See abstract.}
  \item Did you specify all the training details (e.g., data splits, hyperparameters, how they were chosen)?
    \answerYes{See appendix~\ref{sec:appendix_implementation}}
        \item Did you report error bars (e.g., with respect to the random seed after running experiments multiple times)?
    \answerNA{}
        \item Did you include the total amount of compute and the type of resources used (e.g., type of GPUs, internal cluster, or cloud provider)?
    \answerYes{See Sec~\ref{sec:ablations}.}
\end{enumerate}

\item If you are using existing assets (e.g., code, data, models) or curating/releasing new assets...
\begin{enumerate}
  \item If your work uses existing assets, did you cite the creators?
    \answerYes{See linked code.}
  \item Did you mention the license of the assets?
    \answerYes{See linked code.}
  \item Did you include any new assets either in the supplemental material or as a URL?
    \answerNo{}
  \item Did you discuss whether and how consent was obtained from people whose data you're using/curating?
    \answerNA{}
  \item Did you discuss whether the data you are using/curating contains personally identifiable information or offensive content?
    \answerNA{}
\end{enumerate}

\item If you used crowdsourcing or conducted research with human subjects...
\begin{enumerate}
  \item Did you include the full text of instructions given to participants and screenshots, if applicable?
    \answerNA{}{}
  \item Did you describe any potential participant risks, with links to Institutional Review Board (IRB) approvals, if applicable?
    \answerNA{}{}
  \item Did you include the estimated hourly wage paid to participants and the total amount spent on participant compensation?
    \answerNA{}
\end{enumerate}

\end{enumerate}


\clearpage
\input{appendix}

\clearpage


\end{document}

%% file: math_commands.tex
\usepackage{amsmath,amsfonts,bm}

\def\eqref#1{equation~\ref{#1}}

\def\1{\bm{1}}

\DeclareMathAlphabet{\mathsfit}{\encodingdefault}{\sfdefault}{m}{sl}
\SetMathAlphabet{\mathsfit}{bold}{\encodingdefault}{\sfdefault}{bx}{n}

\DeclareMathOperator*{\argmax}{arg\,max}
\DeclareMathOperator*{\argmin}{arg\,min}

\newcommand{\Tau}{\mathcal{T}}

%% file: abstract.tex
\begin{abstract}
\label{abstract}
While behavior learning has made impressive progress in recent times, it lags behind computer vision and natural language processing due to its inability to leverage large, human-generated datasets.
Human behaviors have wide variance, multiple modes, and human demonstrations typically do not come with reward labels.
These properties limit the applicability of current methods in Offline RL and Behavioral Cloning to learn from large, pre-collected datasets.
In this work, we present Behavior Transformer (BeT), a new technique to model unlabeled demonstration data with multiple modes. 
BeT retrofits standard transformer architectures with action discretization coupled with a multi-task action correction inspired by offset prediction in object detection.
This allows us to leverage the multi-modal modeling ability of modern transformers to predict multi-modal continuous actions.  
We experimentally evaluate BeT on a variety of robotic manipulation and self-driving behavior datasets.
We show that BeT significantly improves over prior state-of-the-art work on solving demonstrated tasks while capturing the major modes present in the pre-collected datasets.
Finally, through an extensive ablation study, we analyze the importance of every crucial component in BeT. 
Videos of behavior generated by BeT are available here: \url{https://mahis.life/bet}.
\end{abstract}

%% file: introduction.tex
\section{Introduction}

Creating agents that can behave intelligently in complex environments has been a longstanding problem in machine learning.
Although Reinforcement Learning~(RL) has made significant advances in behavior learning, its success comes at the cost of high sample complexity~\cite{mnih2015human,duan2016benchmarking,akkaya2019solving}. Without priors on how to behave, state-of-the-art RL methods require online interactions on the order of 1-10M `reward-labeled' samples for benchmark control tasks~\cite{yarats2021mastering}. This is in stark contrast to vision and language tasks, where pretrained models and data-driven priors are the norm~\cite{devlin2018bert,brown2020language,byol,bardes2021vicreg}, which allows for efficient downstream task solving.

\begin{figure}
    \centering
    \includegraphics[width=1.0\linewidth]{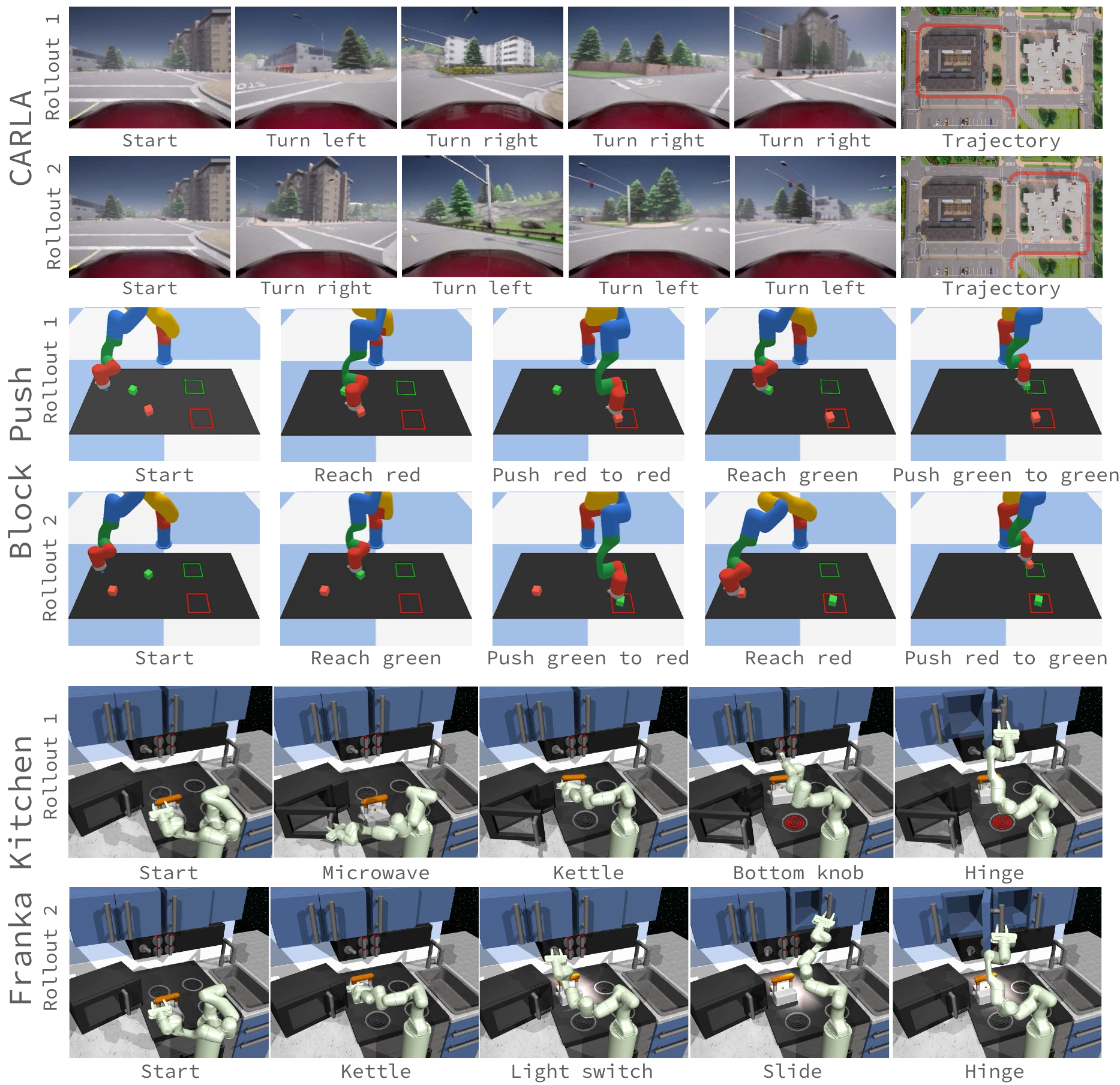}
    \caption{Unconditional rollouts from \method{} models trained from multi-modal demonstartions on the CARLA, Block push, and Franka Kitchen environments. Due to the multi-modal architecture of \method{}, even in the same environment successive rollouts can achieve different goals or the same goals in different ways. 
    }
    \label{fig:intro_rollouts}
    \vspace{-25pt}
\end{figure}

So how do we learn behavioral priors from pre-collected data? One option is offline RL~\cite{levine2020offline}, where offline datasets coupled with conservative policy optimization can learn task-specific behaviors. However, such methods have yet to tackle domains where task-specific reward labels are not present. Without explicit reward labels, imitation learning, particularly behavior cloning, is a more fitting option~\cite{pomerleau1989alvinn,bojarski2016end,torabi2018behavioral}. Here, given behavior data $\mathcal{D} \equiv \{s_t,a_t\}$, behavior models can be trained to predict actions $f_{\theta}(s_t) \rightarrow a_t$ through supervised learning. When demonstration data is plentiful, such approaches have found impressive success in a variety of domains from self-driving~\cite{pomerleau1989alvinn,codevilla2019exploring} to robotic manipulation~\cite{zhang2018deep,pari2021surprising}. Importantly, it requires neither online interactions nor reward labels. 

However, state-of-the-art behavior cloning methods often make a fundamental assumption -- that the data is drawn from a unimodal expert solving a single task. This assumption is often baked in to the architecture design, such as using a Gaussian prior. On the other hand, natural pre-collected data is sub-optimal, noisy, and contains multiple modes of behavior, all entangled in a single dataset. This distributionally multi-modal experience is most prominent in human demonstrations. Not only do we perform a large variety of behaviors every day, our personal biases result in significant multi-modality even for the same behavior~\cite{grauman2021ego4d,lynch2020learning}. Current approach for behavior cloning from such datasets primarily focus on learning goal-conditioned policies, where each goal implies a single mode of behavior~\citep{hausman2017multi,gupta2019relay,lynch2020learning,dasari2020transformers}. 
However, even after goal-conditioning, an important question remains: How do we train models that can natively ``clone'' multi-modal behavior data? 

In this work, we present Behavior Transformers (\method{}), a new method for learning behaviors from rich, distributionally multi-modal data. 
\method{} is based of three key insights. 
First, we leverage the context based multi-token prediction ability of transformer-based sequence models~\cite{vaswani2017attention} to predict multi-modal actions. 
Second, since transformer-based sequence models are naturally suited to predicting discrete classes, we cluster continuous actions into discrete bins using k-means~\cite{macqueen1967some}.
This allows us to model high-dimensional, continuous multi-modal action distributions as categorical distributions without learning complicated generative models~\cite{kingma2013auto,dinh2016density}. 
Third, to ensure that the actions sampled from \method{} are useful for online rollouts, we concurrently learn a residual action corrector to produce continuous actions for a sampled action bin.

We experimentally evaluate \method{} on five datasets ranging from simple diagnostic toy datasets to complex datasets that include simulated robotic pushing~\cite{florence2021implicit}, sequential task solving in kitchen environments~\cite{gupta2019relay}, and self-driving with visual observations in CARLA~\cite{carla}. The two main findings from these experiments can be summarized as:
\begin{enumerate}[leftmargin=0.05\textwidth]
    \item On multi-modal datasets, \method{} achieves significantly higher performance during online rollouts compared to prior behavior modelling methods.
    \item Rather than collapsing or latching onto one mode, \method{} is able to cover the major modes present in the training behavior datasets. Unconditional rollouts from this model can be seen in Fig.~\ref{fig:intro_rollouts}.
\end{enumerate} 

All of our datasets, code, and trained models will be made publicly available.

%% file: method.tex
\section{Behavior Transformers}

Given a dataset of continuous observation and action pairs $\mathcal D \equiv \{(o, a)\} \subset \mathcal O \times \mathcal A$ that contains behaviors we are interested in, our goal is to learn a behavior policy $\pi : \mathcal O \mapsto \mathcal A$ that models this data without any online interactions with the environment or reward labels.
This setup follows the Behavior Cloning formulation, where policies are trained to model demonstrations from expert rollouts.
Often, such policies are chosen from a hypothesis class parametrized by parameter set $\theta$.
Following this convention, our objective is to find the parameter $\theta$ that maximizes the probability of the observed data
\begin{equation}
    \theta^* := \argmax_\theta \prod_t \mathbb P(a_t \mid o_t; \theta)
\end{equation}
When the model class is restricted to unimodal isotropic Gaussians, this maximum likelihood estimation problem leads to minimizing the Mean Squared Error (MSE), $\sum_t \| a_t - \pi(o_t; \theta) \|^2$. 

\paragraph{Limitations of traditional MSE-based BC:}
\begin{wrapfigure}{r}{0.6\textwidth}
\vspace{-10pt}
    \centering
    \includegraphics[width=0.58\textwidth]{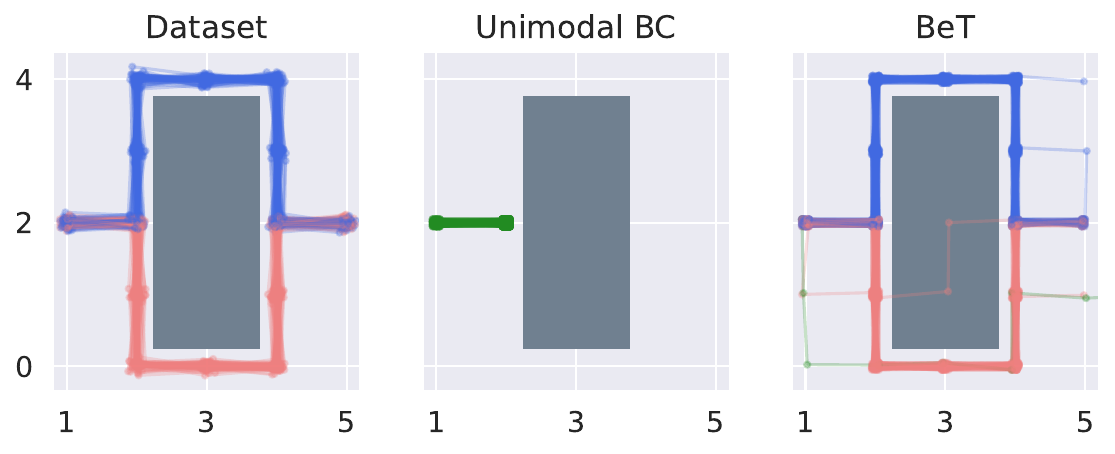}
    \caption{Comparison between a regular MSE-based BC model and a \method{} models that can capture multi-modal distributions. The MSE-BC model takes $0$ action to minimize MSE.}
    \label{fig:multipath_intro}
\vspace{-5pt}
\end{wrapfigure}
While MSE-based BC has been able to solve a variety of tasks \cite{bojarski2016end,torabi2018behavioral}, it assumes that the data distribution is unimodal.
Clean data from an expert demonstrator solving a particular task in a particular way satisfies this assumption, but pre-collected intelligent behavior often may not~\cite{lynch2020learning, gupta2019relay}.
While more recent behavior generation models have sought to address this problem, they often require complex generative models \citep{singh2020parrot}, an exponential number of bins for actions \citep{mandi2021towards}, complicated training schemes \cite{spirl}, or time-consuming test-time optimization \citep{florence2021implicit}.
An experimental analysis of some of these prior works is presented in Section~\ref{sec:results}.

\begin{figure}[t!]
    \centering
    \includegraphics[width=\linewidth]{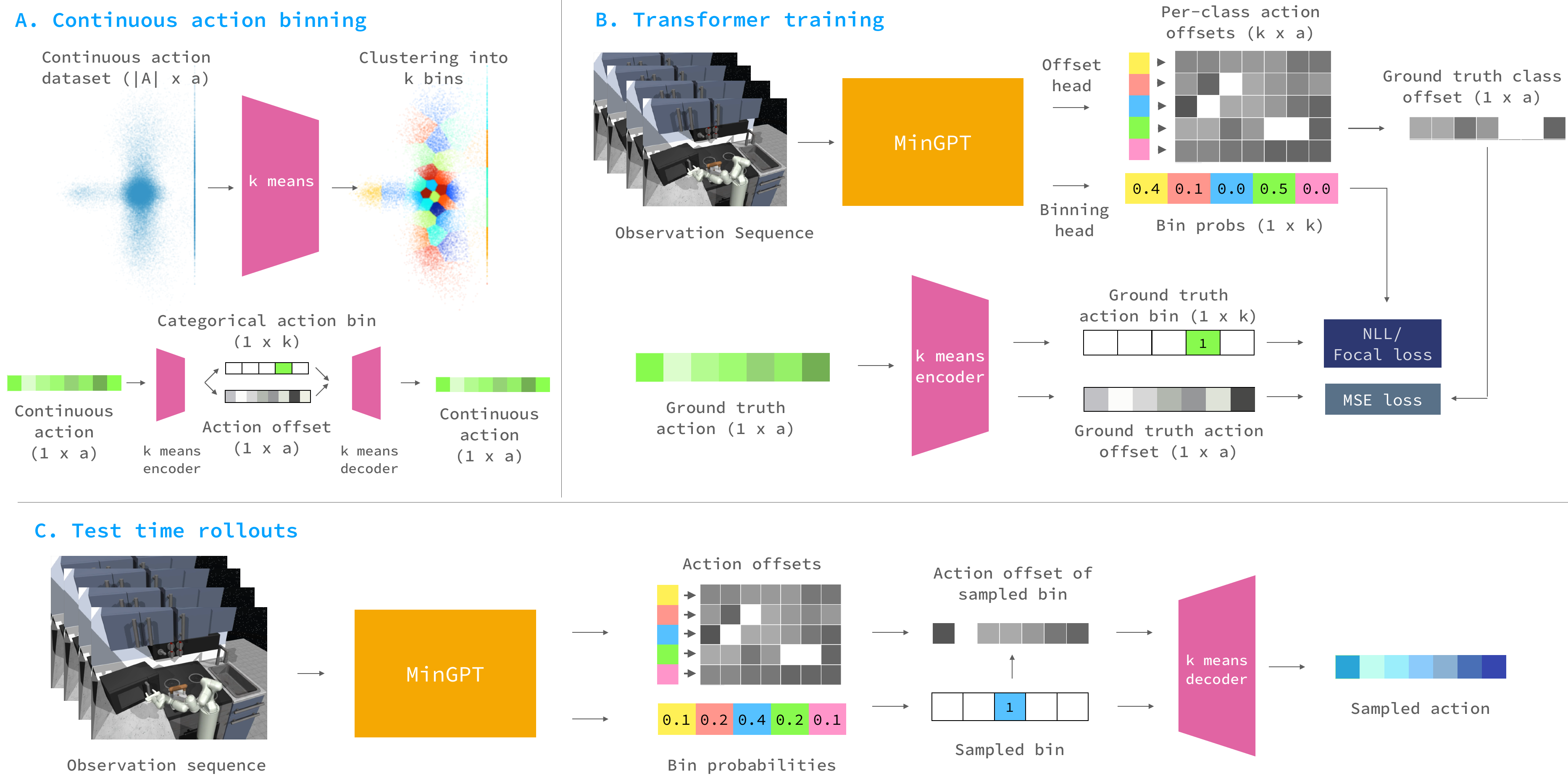}
    \caption{Architecture of Behavior Transformer. (A) The continuous action binning using k-means algorithm that lets \method{} split every action into a discrete bin and a continuous offset, and later combine them into one full action. (B) Training \method{} using demonstrations offline; each ground truth action provides a ground truth bin and residual action, which is used to train the minGPT trunk with its binning and action offset heads. (C) Rollouts from \method{} in test time, where it first chooses a bin and then picks the corresponding offset to reconstruct a continuous action.}
    \label{fig:arch}
\end{figure}

\paragraph{Overview of Behavior Transformers (\method{}):}
We address two critical assumptions in regular BC.
First, we relax the assumption that the behavior we are cloning is purely Markovian, and instead model $P(a_t \mid o_t, o_{t-1}, \cdots, o_{t-h+1})$ for some horizon $h$.
Second, instead of assuming that actions are generated by a unimodal action distribution, we model our action distribution as a mixture of gaussians.
However, unlike previous efforts similar to Mixture Density Networks (MDN) to do so, whose limitations have been explored in \citet{florence2021implicit}, we do not explicitly predict mode centers, which significantly improves our modeling capacity.
To operationalize these two features in a single behavior model, we make use of transformers since (a) they are effective in utilizing prior observational history, and (b) they are naturally suited to output multi-modal tokens through their architecture.

\subsection{Action discretization for distribution learning}
Although transformers have become standard as a backbone for sequence-to-sequence models~\cite{devlin2018bert,brown2020language}, they are designed to process discrete tokens and not continuous values.
In fact, modeling multi-modal distributions of high-dimensional continuous variables in a tractable manner is in itself a challenging problem, especially if we want the trained behavior model to cover the modes present in the dataset.
To address this, we propose a new factoring of the action prediction task by dividing each action in two parts: a categorical variable denoting an `action center', and a corresponding `residual action'.

To this end, given the actions in our dataset, we first optimize for a set of $k$ action centers, $\{A_1, A_2, \cdots, A_k\} \subset \mathcal A$.
We then decompose each action into two parts: a categorical variable representing the closest action bin, $\lfloor a \rfloor := \argmin_{i} \|a - A_i\|_2$, and a continuous residual action $\langle a \rangle := a - A_{\lfloor a \rfloor}$. 
If we are given the set of action centers $\{A_i\}_{i=1}^k$, an action bin index $\lfloor a \rfloor$ and the residual action $\langle a \rangle$, we can deterministically reconstruct the true action $a := A_{\lfloor a \rfloor} + \langle a \rangle$.
Once learned, these k-means based encoder and decoders for this action factorization process are fixed for the rest of the train and testing phases.
The action factorization procedure is illustrated in Fig.~\ref{fig:arch}~(A).

\subsection{Attention-based behavior mode learning}
Once we have the clustering based autoencoder learned from the actions in the dataset, we model our demonstration trajectories with \method{}.
We use a transformer decoder model, namely minGPT~\cite{brown2020language}, with minor modifications, as our backbone.
The transformer $\Tau$ takes in a sequence of continuous observations $(o_i, o_{i+1}, \cdots, o_{i+h-1})$ and learns a sequence-to-sequence model mapping each observation to a categorical distribution over $k$ discrete action bins.
The predicted probability sequence is then compared with the ground truth labels, $(\lfloor a_i \rfloor, \lfloor a_{i+1} \rfloor, \lfloor a_{i+2} \rfloor, \cdots, \lfloor a_{i+h-1} \rfloor)$.
We use a negative log-likelihood-based Focal loss~\cite{lin2017focal} between the predicted categorical distribution probabilities and the ground truth labels to train the transformer head. 
Focal loss is a simple modification over the standard cross entropy loss. While the standard cross entropy loss for binary classification can be thought of $\mathcal L_{ce} (p_t) = -\log (p_t)$, Focal loss adds a term $(1-p_t)^\gamma$ to this, to make the new loss
\[\mathcal L_{focal} (p_t) = -(1-p_t)^\gamma\log (p_t)\]
This loss has the interesting property that its gradient is more steep for smaller values of $p_t$, while flatter for larger values of $p_t$.
Thus, it penalizes and changes the model more for making errors in the low-probability classes, while is more lenient about making errors in the high probability classes.
The model is illustrated in Fig.~\ref{fig:arch}~(B). 

\subsection{Action correction: from coarse to finer-grained predictions}

Using a transformer allows us to model multi-modal actions. However, discretizing the continuous action space in any way invariably causes loss of fidelity~\cite{janner2021offline}. Discretization error may cause online rollouts of the behavior policy to go out of distribution from the original dataset~\cite{dagger2010}, which can in turn cause critical failures.
To predict the complete continuous action, we add an extra head to the transformer decoder that offsets the discretized action centers based on the observations.

For each observation $o_i$ in the sequence, the head produces a $k \times \text{dim}(A)$ matrix with $k$ proposed residual action vectors, $\left (\langle a^{(j)}_i \rangle \right )_{j=1}^k = (\langle \hat a^{(1)}_i \rangle, \langle \hat a^{(2)}_i \rangle, \langle\hat a^{(3)}_i \rangle, \cdots, \langle\hat a^{(k)}_i \rangle)$, where the residual actions correspond to bin centers $A_1, A_2, A_3,  \cdots, A_k$.
These residual actions are trained with a loss akin to the \textit{masked multi-task loss}~\cite{girshick2015fast} from object detection. 
In our case, if the ground truth action is $\mathbf{a}$, the loss is:
\begin{equation}
    \text{MT-Loss}\left (\mathbf{a}, \left (\langle \hat a^{(j)}_i \rangle\right )_{j=1}^k\right ) = \sum_{j=1}^k \mathbb I [\lfloor \mathbf{a} \rfloor = j] \cdot \| \langle \mathbf{a} \rangle - \langle \hat a^{(j)} \rangle \|_2^2
\end{equation}
Where $\mathbb I[]$ denotes the Iverson bracket, ensuring the offset head of \method{} only incurs loss from the ground truth class of action $\mathbf{a}$.
This mechanism prevents the model from trying to fit the ground truth action using the offset at every index.

\subsection{Test-time sampling from \method{}}
During test time, at timestep $t$ we input the latest $h$ observations $(o_t, o_{t-1}, \cdots, o_{t-h+1})$ to the transformer, combining the present observation $o_t$ with $h-1$ previous observations.
Our trained MinGPT model gives us $h \times 1 \times k$ bin center probability vectors, and $h \times k \times \text{dim}(A)$ offset matrix.
To sample an action at timestep $t$, we first sample an action center according to the predicted bin center probabilities on the $t$\textsuperscript{th} index.
Once we have chosen an action center $A_{t,j}$, we add the corresponding residual action $\langle \hat a_t^{(j)} \rangle$ to it to recover a predicted continuous action 
$\hat {\mathbf{a_t}} = A_{t,j} + \langle \hat a_t^{(j)} \rangle$. This sampling procedure is illustrated in Fig.~\ref{fig:arch}~(C).

%% file: results.tex
\section{Experiments}
\label{sec:exp}
\label{sec:results}

We now study the empirical performance of \method{} on a variety of behavior learning tasks. Our experiments are designed to answer the following questions: (a) Is \method{} able to imitate multi-modal demonstrations? (b) How well does \method{} capture the modes present in behavior data? (c) How important are the individual components of \method{}? 

\subsection{Environments and datasets}
We experiment with five broad environments. While full descriptions of these environments, dataset creation procedure, and overall statistics are in Appendix~\ref{sec:appendix_environments_datasets}, a brief description of them are as follows. 

\begin{enumerate}[label=(\alph*),leftmargin=*]
    \item \textbf{Point mass environment \#1:} Our first set of experiments in Fig.~\ref{fig:multipath_intro}, used to get a qualitative understanding of \method{}, were performed in a simple Pointmass environment with a 2D observation and action space with two hundred demonstrations. The pre-collected demonstrations start at a fixed point, and then make their way to another point while avoiding a block in the middle. The two primary modes in this dataset are taking a left turn versus a right turn.
    
    \item \textbf{Point mass environment \#2:} The setup is similar to the previous environment with the exception of one straight line and two complicated prolonged `Z' shaped modes of demonstration (Fig.~\ref{fig:multipath_time}.)
    
    \item \textbf{CARLA self-driving environment:} CARLA~\cite{carla} uses the Unreal Engine to provide a simulated driving environment in a visually realistic landscape. The agent action space is 2D (accelerate/brake and left/right steer), while the observation space is (224,224,3)-dimensional RGB image from the car. A hundred total demonstrations drive around a building block in two distinct modes. This environment highlights the challenge of behavior learning from high-dimensional observations as shown in Fig.~\ref{fig:intro_rollouts} (a). For visual observations with \method{}, we use a frozen ResNet-18~\cite{he2016deep} pretrained on ImageNet~\cite{deng2009imagenet} as an encoder.

    \item \textbf{Multi-modal block-pushing environment:} For more complicated interaction data, we use the multi-modal block-pushing environment from Implicit Behavioral Cloning (IBC)~\cite{florence2021implicit}, where an XArm robot needs to push two blocks into two squares in any order. The blocks and target squares are colored red and green. The positions of the blocks are randomized at episode start. We collect 1,000 demonstrations using a deterministic controller with two independent axes of multi-modality: (a) it starts by reaching for either the red or the green block, with 50\% probability, and (b) it pushes the blocks to (red, green) or (green, red) squares respectively with 50\% probability.
    
    \item \textbf{Franka kitchen environment:} To highlight the complexity of performing long sequences of actions, we use the Relay Kitchen Environment \cite{gupta2019relay} where a Franka robot manipulates a virtual kitchen environment.
    We use the relay policy learning dataset with 566 demonstrations collected by human participants wearing VR headsets. The participants completed a sequence of four object-interaction tasks in each episode~\cite{gupta2019relay}.
    There are a total of seven interactable objects in the kitchen: a microwave, a kettle, a slide cabinet, a hinge cabinet, a light switch, and two burner knobs. This dataset contains two different kinds of multi-modality: one from the inherent noise in human demonstrations, and another from the demonstrators' intent.
\end{enumerate}

\subsection{Baseline behavior learning methods}
\looseness=-1 While a full description of our baselines are in Appendix~\ref{sec:app:baselines}, a brief description of them is here:

\begin{enumerate}[label=(\alph*),leftmargin=*]
    \item \textbf{Multi-layer Perceptron with MSE (RBC):} We use MLP networks trained with MSE loss as our first baseline, since this is the standard way of performing behavioral cloning for a new task~\cite{torabi2018behavioral}. A comparison with transformer-based behavior cloning is discussed in Section~\ref{sec:ablations}.
    \item \textbf{Nearest neighbor (NN):} Nearest neighbor based algorithms are easy to implement, and has recently shown to have strong performance on complicated behavioral cloning tasks~\cite{arunachalam2022dexterous}.  
    \item \textbf{Locally Weighted Regression (LWR):} This non-parametric approach provides better regularization compared to NN and is a strong alternative to parametric BC~\cite{atkeson1997locally,pari2021surprising}.
    \item \textbf{Variational auto-encoders (VAE):} Inspired by SPiRL~\cite{spirl}, where behavioral priors are learned through a VAE~\cite{kingma2013auto}, we compare with continuous actions generated from the VAE and the prior. 
    \item \textbf{Normalizing Flow (Flow):} Inspired by PARROT~\cite{singh2020parrot}, where state-conditioned action priors are learned through a Flow model~\cite{dinh2016density}, we compare with actions generated from the Flow model. 
    \item \textbf{Implicit Behavioral Cloning (IBC):} Instead of modeling the conditional distribution $P(a \mid o)$, IBC models the joint probability distribution $P(a, o)$ using energy-based models~\cite{florence2021implicit}. While IBC is slower than explicit BC models because of their sampling requirements, they have been shown to learn well on multi-modal data, and outperform earlier work such as MLP-MDNs~\cite{bishop1994mixture}. 
\end{enumerate}

\subsection{Is \method{} able to imitate multi-modal demonstrations?}
The first question we ask is whether \method{} can actually clone behaviors given a mixed dataset of unlabeled, multi-modal behaviors.
To examine that, we look at the performance of our model in CARLA, Block push, and Kitchen environments compared with our baselines in Table~\ref{tab:perf-table}.

\input{tables/perf_table}

We see that \method{} outperforms all other methods in all environments except CARLA, where it is narrowly outperformed by LWR. 
Since the models are all behavioral cloning algorithms, they share the failure mode of failing once the observations go out of distribution (OOD).
However, they vary in the tolerance.
For example, \method{} shines in the Block push environment, where alongside extreme environment randomness and multi-modality, the models also have to learn significant long-term behaviors and commit to a single mode over a long period.
While all baselines can somewhat successfully reach one block, they fail to complete the long-horizon, multi-modal task of pushing two blocks into two different bins.
On the other hand, we observe that \method{}'s primary failure mode is not realizing a block has not completely entered the target yet, while other methods either go OOD quickly, or keep switching between modes.
We also observe that \method{} performs well even in complex observation and action spaces.
In the CARLA environment, the model takes in visual observations, while in the Franka Kitchen environment, the action space corresponds to a 9-DOF torque controlled robot.
BeT handles both cases with the same ease as it does environments with lower-dimensional observation or action spaces.

\subsection{Does \method{} capture the modes present in behavior data?}

\begin{figure}
    \centering
    \includegraphics[width=\linewidth]{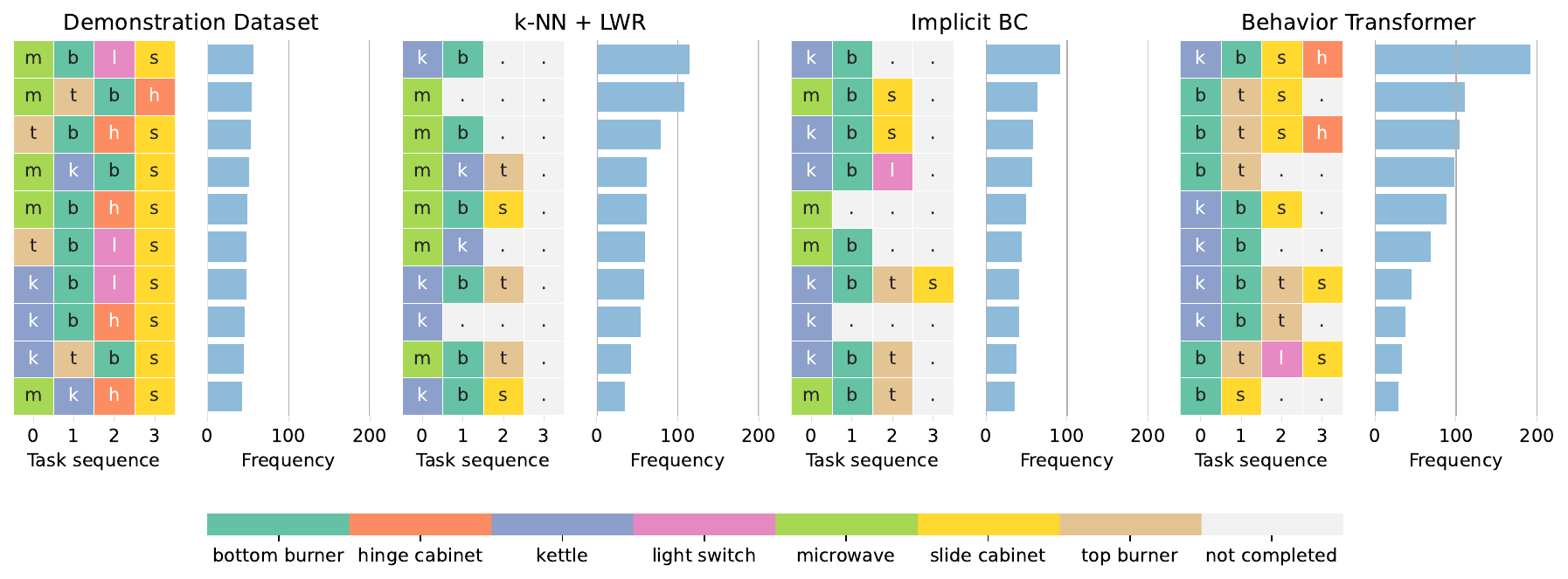}
    \caption{Distribution of most frequent tasks completed in sequence in the Kitchen environment. Each task is colored differently, and frequency is shown out of a 1,000 unconditional rollouts from the models.}
    \label{fig:multimodal_kitchen_tasks}
\end{figure}
Next, we examine the question of whether, given a dataset where multi-modal behavior exists, our model learns behavior that is also multi-modal.
Here, we are interested in seeing the variance of the behavior of the model over different rollouts.
In each of our environments, the demonstrations contain different types of multi-modality.
As a result, we show a comprehensive analysis of multi-modality seen in our agent behaviors.

\input{tables/multimodality_table}

We see in Table~\ref{tab:multimodal-table} that in CARLA and Block push, \method{} covers all the modes of the demonstration data, even in the few cases where it does not perfectly match the demonstrated task probabilities.
For the Kitchen environment, we see in Fig.~\ref{fig:multimodal_kitchen_tasks} that \method{} visits certain strings of tasks more frequently than in the original demonstrations.
However, compared to other strong baselines, \method{} generates longer task strings more often while maintaining diversity and not collapsing to a single mode.

\subsection{How important are the individual components of \method{}?}
\label{sec:ablations}
There are four key differences between \method{} architecture and standard BC: (a) binning actions into discrete clusters, (b) using offsets to faithfully reconstruct actions later, (c) learning sequentially to use historical context, and (d) using an attention-based MinGPT trunk.
In this section, we discuss the impacts they have in \method{} performance.

\input{tables/ablation_table}

\paragraph{Impact of discrete binning:}
Intuitively, having discrete options for bin centers is what enables \method{} to express multi-modal behavior even when starting from an identical starting state.
Indeed, if there is no binning, we see from Table~\ref{tab:ablation-table} that the performance of \method{} drops significantly.
More tellingly, in the Franka Kitchen environment, the model only ever completed a subsequence of (kettle, top/bottom burner, light switch, slide cabinet) tasks after 100 random rollouts.
This result shows us that having discrete bins helps \method{} achieve multi-modality.
We also experiment with the Mixture density networks (MDN) \cite{bishop1994mixture} and uniform quantization, as shown in previous works \cite{florence2021implicit, janner2021offline}.
We see that they may perform well sometimes but overall still fall short of our k-means binning approach.

\paragraph{Necessity of action offsets:}
An important feature of \method{} is the residual action offset that corrects the discrete actions coming from the bins.
While the bin centers may be quite expressive, Table~\ref{tab:ablation-table} shows that the inability to correct them causes a performance degrade.
Interestingly, the largest degradation comes in the Kitchen environment, which also has the highest dimensional action space.
Intuitively, we can understand how in higher dimension the loss of fidelity from discretizing would be higher, and the relative performance loss across three environments support that hypothesis.

\paragraph{Importance of historical context:}
\begin{figure}
    \centering
    \includegraphics[width=\linewidth]{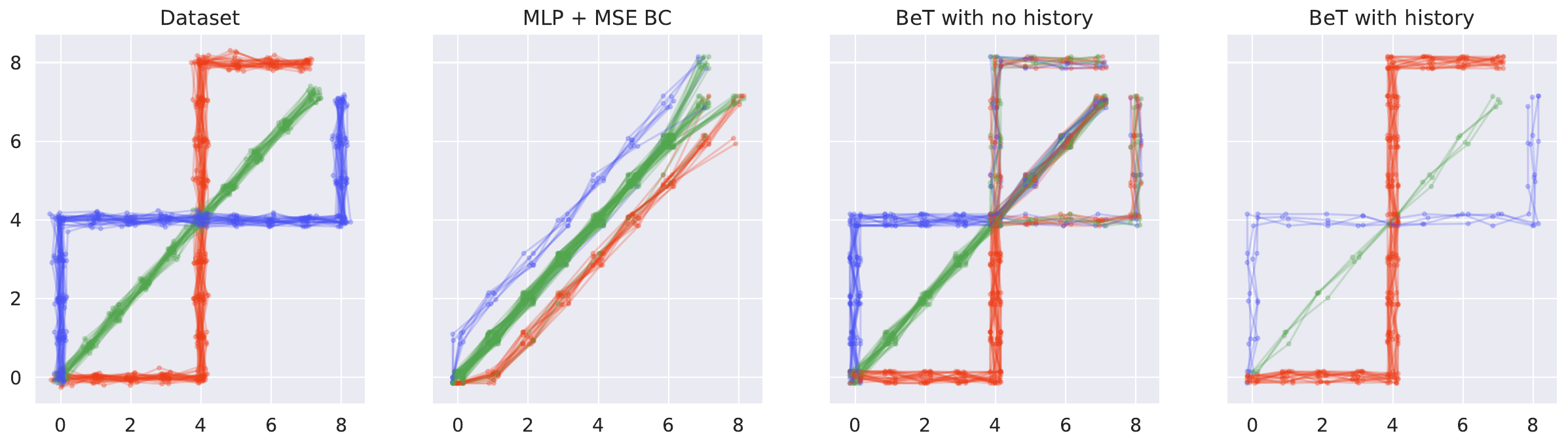}
    \caption{Comparison between an RBC model and two \method{} models, trained with and without historical context on a dataset with three distinct modes. \method{} with history is better able to capture the context-dependant behavior in the demonstrations.}
    \label{fig:multipath_time}
    \vspace{-15pt}
\end{figure}

While RL algorithms traditionally assume environments are Markovian, human behavior in an open-ended environment is rarely so.
Thus, using historical context helps \method{} to perform well.
We show a simple experiment in Fig.~\ref{fig:multipath_time} on the second point mass environment.
Here, training and evaluating with some historical context allows \method{} to follow the demonstrations better.
We experience the same in the CARLA, Block push, and Kitchen environments, where training with some historical context raises performance across the board as seen in Table~\ref{tab:ablation-table}.

\paragraph{Importance of transformer architecture:}
Despite transformers' success in other fields of machine learning, it is natural to wonder whether the tasks \method{} solves here really requires one.
We ablated \method{} by replacing the MinGPT trunk with an MLP, Temporal Convolution, and LSTMs, and found that they have lower performance while also being difficult to train stably. This performance reduction remains even if the MLP is given some historical context by stacking $h$ observations before passing it to the MLP. See Table.~\ref{tab:ablation-table} for results and Appendix ~\ref{sec:app:ablation:transformer} for further details.

\paragraph{Computation considerations:}
While transformers in usual contexts are large models, we downscale them for our application in \method{} (See Appendix~\ref{sec:appendix_arch}). Our models contain on the order of $10^4$-$10^6$ parameters, and even with a small batch size trains within an hour for our largest datasets (Block push) on a single desktop GPU. In contrast, for the same task, our strongest baseline IBC takes about 14 hours. Evaluation rollouts on the same environment take 1.65 seconds with \method{}, as opposed to 17.70 seconds with IBC.

%% file: tables/perf_table.tex
\begin{table}
\centering
\caption{Performance of \method{} compared with different baselines in learning from demonstrations. For CARLA, we measure the probability of the car reaching the goal successfully. For Block push, we measure the probability of reaching one and two blocks, and the probabilities of pushing one and two blocks to respective squares. For Kitchen, we measure the probability of $n$ tasks being completed by the model within the allotted  280 timesteps. Evaluations are over 100 rollouts in CARLA and 1,000 rollouts in Block push and Kitchen environments.}
\label{tab:perf-table}
\begin{tabular}{lcccccccccc}
          & CARLA      & \multicolumn{4}{c}{Block push}                             & \multicolumn{5}{c}{Kitchen}                                                \\ \cline{2-11} 
          & Driving    & \multicolumn{2}{c}{Reach}  & \multicolumn{2}{c}{Push}      & \multicolumn{5}{c}{\# Tasks completed}                                        \\
Baselines & Success    & R1         & R2            & P1            & P2            & 1          & 2             & 3             & 4             & 5             \\ \hline
RBC       & 0.98       & 0.67       & 0             & 0             & 0             & 0          & 0             & 0             & 0             & 0             \\
1-NN      & 0.99       & 0.49       & 0.05          & 0.01          & 0             & 0.90       & 0.72          & 0.44          & 0.17          & 0             \\
LWR       & \textbf{1} & 0.50       & 0.06          & 0             & 0             & \textbf{1} & 0.83          & 0.52          & 0.21          & 0             \\
VAE       & 0          & 0.60       & 0.05          & 0             & 0             & \textbf{1} & 0             & 0             & 0             & 0             \\
Flow      & 0.03       & 0.59       & 0.02          & 0             & 0             & 0.04       & 0             & 0             & 0             & 0             \\
IBC       & 0.25       & 0.98       & 0.04          & 0.01          & 0             & 0.99       & 0.87          & 0.61          & 0.24          & 0             \\
BeT (Ours)      & 0.98       & \textbf{1} & \textbf{0.99} & \textbf{0.96} & \textbf{0.71} & 0.99       & \textbf{0.93} & \textbf{0.71} & \textbf{0.44} & \textbf{0.02}
\end{tabular}
\end{table}

%% file: tables/multimodality_table.tex
\begin{table}
    \caption{Multimodality learned from the multimodal demonstrations by different algorithms. In CARLA, we consider the probability of turning left vs. right at the intersection, ignoring OOD rollouts. In Block push, we consider two set of probabilities, (a) which block was reached first, and (b) what was the pushing target for each block. Finally, in Franka Kitchen, we consider the empirical entropy for the task sequences, considered as strings, sampled from the model. We highlight the values closest to the corresponding demonstration values.}
\label{tab:multimodal-table}
\begin{tabular}{lccccccccc}
               & \multicolumn{2}{c}{CARLA} & \multicolumn{2}{c}{\begin{tabular}[c]{@{}c@{}}Block: first\\ block reached\end{tabular}} & \multicolumn{2}{c}{\begin{tabular}[c]{@{}c@{}}Push: red\\ block target\end{tabular}} & \multicolumn{2}{c}{\begin{tabular}[c]{@{}c@{}}Push: green\\ block target\end{tabular}} & \begin{tabular}[c]{@{}c@{}}Kitchen\\ \end{tabular} \\ \cline{2-10} 
Baselines      & Left        & Right       & Red                                       & Green                                     & Red                                      & Green                                    & Red                                       & Green                                     & Task entropy                                                        \\ \hline
RBC            & 0           & 0.98        & 0.41                                      & 0.25                                      & 0                                        & 0                                        & 0                                         & 0                                         & 0                                                            \\
1-NN           & 0           & 0.99        & 0.24                                      & 0.25                                      & 0                                     & 0                                     & 0                                      & 0.01                                      & 2.12                                                            \\
LWR            & 0           & 1           & 0.26                                      & 0.26                                      & 0.01                                     & 0                                     & 0.01                                      & 0.01                                      & 2.29                                                            \\
VAE            & 0           & 0           & 0.27                                      & 0.33                                      & 0                                        & 0                                        & 0                                         & 0                                         & 0.72                                                            \\
Flow           & 0           & 0           & 0.31                                      & 0.29                                      & 0                                     & 0                                     & 0                                         & 0                                         & 0.08                                                            \\
IBC            & 0.12        & 0.13        & \textbf{0.48}                                      & \textbf{0.50}                                      & 0                                        & 0                                        & 0.01                                      & 0.01                                     & 2.41                                                            \\
BeT (Ours)           & \textbf{0.34}        & \textbf{0.64}        & 0.54                                      & 0.46                                      & \textbf{0.43}                                     &\textbf{ 0.44}                                     & \textbf{0.41}                                      & \textbf{0.40 }                                     & \textbf{2.47}                                                           \\ \hline
Demonstrations & 0.50        & 0.50        & 0.50                                      & 0.50                                      & 0.50                                     & 0.50                                     & 0.50                                      & 0.50                                      & 2.96                                                              
\end{tabular}
\end{table}

%% file: tables/ablation_table.tex
\begin{wraptable}{r}{70mm}
    \vspace{-15pt}
    \caption{Relative performance of ablated variants of BeT, normalized by average BeT successes at the task}
    \label{tab:ablation-table}
    \centering
    \begin{tabular}{cccc}
    Ablations  & CARLA & Block push & Kitchen \\ \hline
    No offsets & 0.94  & 0.95       & 0.78    \\
    No binning & 0.94  & 0.25       & 0.68       \\
    No history & 0.65  & 0.95       & 0.88   \\
    MLP     & 0.90  & 0       & 0.05  \\
    Temp. Conv    & 0.72  & 0.01       & 0.26 \\
    LSTM     & 0.03  & 0.03       & 0.04  \\
    GPT-MDN     & 0.30  & 0.83       & 0.86  \\
    Unif. quant.     & 0.90  & 0.96       & 0.90  
    \end{tabular}
    \vspace{-5pt}
\end{wraptable}

%% file: related_work.tex
\section{Related Work}
This paper builds upon a rich literature in imitation learning, offline learning, generative models, and transformer architectures. The most relevant ones to our work are discussed here.
\paragraph{Learning from offline data:}
Since \citet{pomerleau1988alvinn} showed the possibility of driving an autonomous vehicle using offline data and a neural network, learning behavior from offline data has been a continuous topic of research for scalable behavior learning \citep{argall2009survey, billard2008survey, schaal1999imitation}.
The approaches can be divided into two broad classes: Offline RL \citep{fujimoto2018addressing, kumar2019stabilizing, kumar2020conservative, wu2019behavior, levine2020offline, fu2020d4rl}, focusing on learning from datasets of a mixed quality that also have reward labels;
and imitation learning \citep{osa2018algorithmic, peng2018deepmimic, peng2021amp, ho2016generative}, focusing on learning behavior from a dataset of expert behavior without reward labels.
\method{} falls under the second category, as it is a behavior cloning model.
Behavior cloning is a form of imitation learning that tries to model the action of the expert given the observation which is often used in real-world applications \citep{zhang2018deep, zhu2018reinforcement, zhang2018deep, rahmatizadeh2018vision, florence2019self, zeng2020transporter}.
As behavior cloning algorithms are generally solving a fully supervised learning problem, they tend to be faster and simpler than reinforcement learning or offline RL algorithms and in some cases show competitive results \cite{fu2020d4rl, gulcehre2020rl}.

\paragraph{Generative models for behavior learning:}
One approach for imitation learning is Inverse Reinforcement Learning or IRL \citep{russell1998learning, ng2000algorithms}, where given expert demonstrations, a model tries to construct the reward function. This reward function is then used to generate desirable behavior. GAIL \citep{ho2016generative}, an IRL algorithm, connects generative adversarial models with imitation learning to construct a model that can generate expert-like behavior. Under this IRL framework, previous works have tried to predict multi-modal, multi-human trajectories \citep{lee2016predicting, ivanovic2018generative}. Similarly, other works have tried Gaussian Processes \citep{rasmussen2010gaussian} for creating dynamical models for human motion \citep{wang2007gaussian}.
Another class of algorithms learn a generative action decoder~\citep{pertsch2020accelerating, lynch2020learning, singh2020parrot} from interaction data to make downstream reinforcement learning faster and easier, which inspired \method{}'s action factorization.
Finally, a class of algorithms, most notably ~\citep{liu2020energy, florence2021implicit, kostrikov2021offline, nachum2021provable} do not directly learn a generative model but instead learn energy based models. 
These energy based models can then be sampled to generate desired behavior.
Since \citep{florence2021implicit} is a BC model capable of multi-modality, we compare against it as a baseline in Sec.~\ref{sec:exp}.

\paragraph{Transformers for control:} With the stellar success of transformer models \citep{vaswani2017attention} in natural language processing \citep{devlin2018bert, brown2020language} and computer vision \citep{dosovitskiy2020image}, there has been significant interest in using transformer models to learn behavior and control.
Among those, \citep{chen2021dt, janner2021offline} applies them to Reinforcement Learning and Offline Reinforcement Learning, respectively, while \citep{clever2021assistive, dasari2020transformers, mandi2021towards} use them for imitation learning.
\citep{dasari2020transformers, mandi2021towards} use transformers mostly to summarize historical visual context, while \citep{clever2021assistive} relies on their long-term extrapolation abilities to collect human-in-the-loop demonstrations more efficiently.
\method{} is inspired by both of these use cases, as we use a transformer to summarize historical context while leveraging its generative abilities.
Architecturally, \method{} is most closely related to the imitation learning variant of \citep{janner2021offline}, with a significant difference that while \citep{janner2021offline} learns the joint state, action distribution, \method{} learns the conditional distribution of action given state, which allows \method{} to tackle much more complicated state spaces.

\paragraph{Datasets for distributionally multi-modal data:} 
Similar to computer vision \citep{deng2009imagenet, lin2014microsoft, liu2018large} and natural language processing \citep{bowman2015large,rajpurkar2016squad}, there has been a recent interest in collecting behavior datasets that may aid in downstream behavior learning. Some of them are labeled with agent goals or rewards for downstream tasks \citep{mandlekar2018roboturk,fu2020d4rl, robomimic2021}, while others are more open ended \citep{gupta2019relay,lynch2020learning,young2021playful} and come without reward or task labels. In our work, we focus towards the latter class.
The lack of labeled goal or reward labels in the second category implies that there is more multi-modality in the action distributions compared to action distributions of goal or reward conditioned datasets, which is the same reason a lot of work learning from multi-modal datasets try to learn a goal-conditioned model~\citep{hausman2017multi, gupta2019relay,lynch2020learning,dasari2020transformers}. 
Finally, the lack of labelling requirements mean that the unlabelled datasets are cheaper to obtain, which should help \method{} scale further in the future.

%% file: conclusions.tex
\section{Discussions}
\label{sec:limit}
In this work, we introduce Behavior Transformers (\method{}), which uses a transformer-decoder based backbone with a discrete action mode predictor coupled with a continuous action offset corrector to model continuous actions sequences from open-ended, multi-modal demonstrations.
While \method{} shows promise, the truly exciting use of it would be to learn diverse behavior from human demonstrations or interactions in the real world.
In parallel, extracting a particular, unimodal behavior policy from \method{} during online interactions, either by distilling the model or by generating the right `prompts' \citep{reynolds2021prompt}, would make \method{} tremendously useful as a prior for online Reinforcement Learning.

%% file: appendix.tex
\appendix

\section*{Appendix}
Diverse, multi-modal behaviors generated by our models on different environment are best experienced and understood in a video.
We invite you to visit \url{https://mahis.life/bet} to see \method{} models in action.
\section{Environment and Dataset Details}
\label{sec:appendix_environments_datasets}
\input{appendix_sections/bet_env_datasets}

\section{Implementation Details and Hyperparameters}
\label{sec:appendix_implementation}

\subsection{Baselines}
\input{appendix_sections/bet_baselines}

\subsection{Algorithm Details}
\input{appendix_sections/bet_compute_details}

\subsection{Pseudocode}
\input{appendix_sections/bet_pseudocode}

\subsection{Architecture and Implementation}
\label{sec:appendix_arch}

For our implementation, we used the MinGPT \cite{mingpt} repository almost as-is.
We modified the input token conversion layer to a linear projection layer to handle our continuous, instead of discrete, inputs.
Apart from that, we followed the MinGPT architecture quite exclusively, with successive attention layers with a number of attention head and embedding dimensions.
Between the layers, we used dropout regularization same as \cite{mingpt}.

For the smallest tasks, like point-mass environments, we used models with approximately $10^4$ parameters, which went up to around $10^6$ for Kitchen environments.

\section{Ablation studies}
\label{sec:appendix_ablation}
\input{appendix_sections/bet_ablations}

\newpage

%% file: appendix_sections/bet_env_datasets.tex
\paragraph{Point mass environments:}
In the point mass environment, we have a simple point-mass agent with two-dimensional observation and action spaces. The observation of the agent denotes the $(x, y)$ position of the agent, while the action sets the immediate $(\Delta x, \Delta y)$ displacement of the agent in the next timestep.

To show the effects of unimodal and multimodal behavioral cloning algorithms more cleanly, we also add a ``snapping" effect to the environment which moves the agent close to the nearest integer coordinates after each step.

We generate random trajectories for each of our Multipath experiment datasets.
\begin{enumerate}
    \item In the first one (Fig.~\ref{fig:multipath_intro}), our dataset has two modes, which are colored differently in the figure based on the path taken at the fork.
    \begin{enumerate}
        \item In the first set of demonstrations, the point mass follows the trajectory $(1, 2), (2, 2), (2, 3), (2, 4), (3, 4), (4, 4), (4, 3), (4, 2), (5, 2).$
        \item In the second set of demonstrations, the point mass follows $(1, 2), (2, 2), (2, 1), (2, 0), (3, 0), (4, 0), (4, 1), (4, 2), (5, 2).$
    \end{enumerate}
    \item For the second Multipath environment (Fig.~\ref{fig:multipath_time}), there are three modes of demonstration, which are colored in the figure according to their first step direction. 
    \begin{enumerate}
        \item In the first set of demonstration, the point mass follows $x=y$ from $(0, 0)$ to $(8, 8)$ with $\sqrt 2$ size step increments.
        \item In the second set of demonstration, the point mass follows straight lines from $(0, 0) \rightarrow (0, 4) \rightarrow (4, 4) \rightarrow (8, 4) \rightarrow (8, 8)$ with step size $1$.
        \item In the third set of demonstration, the point mass follows straight lines from $(0, 0) \rightarrow (4, 0) \rightarrow (4, 4) \rightarrow (4, 8) \rightarrow (8, 8)$ with step size $1$.
    \end{enumerate}
\end{enumerate}

\paragraph{CARLA environment:}
We use the CARLA~\cite{carla} self-driving environment to examine \method{} performance in environments with high-dimensional observation spaces. CARLA uses the Unreal Engine to provide a photo-realistic driving simulation.
We create our environment on the Town04 map in CARLA 0.9.13.
The observation space is $224 \times 224 \times 3$ RGB images from the vehicle, which are processed by an ImageNet-pretrained, frozen ResNet-18 to a 512-dimensional real-valued vector.
The action space is $[-1, 1]^2$ with an accelerator-brake axis and a steering axis.

The dataset on this environment is collected with the built-in PID agent with minor tuning.
We fix waypoints in the trajectory that the demonstration agent needs to follow.
The waypoints fork around two central blocks: one set of trajectories thus go to the left, while another set of demonstration trajectories go to the right.
While collecting the demonstrations, we add some noise in the environment before executing an action so that there is some variation in the set of 100 total demonstrations that we collect in the environment.

We do not introduce any traffic participants in this environment intentionally as we intend to show the effects of cleanly bi-modal distributions on the learning algorithms in an environment more complicated than the point-mass environments.

\paragraph{Block-push environment:}

We use a simulated environment similar to Multimodal Push environment described in \cite{florence2021implicit}. We take the environment implementation directly from the PyBullet \cite{coumans2016pybullet} based implementation provided by \citet{florence2021implicit} in \url{https://github.com/google-research/ibc/tree/master/environments}.

In our environment, an XArm robot is situated in front of two blocks in a $0.75 \times 1$ plane. On the plane there are also two square targets. The goal of the agent is to push the blocks inside of the squares. However, the exact order of the block being pushed, or the combination of which block is pushed in which square doesn't matter. A block is considered successfully pushed if the center of the block is less than $0.05$ away from a square.

On initialization, the blocks' positions are randomly shifted within a rectangle of side lengths $(0.2, 0.3)$, while the squares are randomly shifted within a rectangle of size $(0.01, 0.015)$. Additionally, the blocks were rotated at an uniformly arbitrary angle, while the target squares were rotated at an angle between $(\frac{\pi}{6}, -\frac{\pi}{6})$.

The demonstrations in this environments were collected with a hard-coded controller. There are two modes of multimodality inherent in the controller generated demonstartions. The controller:
\begin{enumerate}
    \item Selects a block to start pushing first,
    \item At the same time, independently chooses a target for the block to be pushed into.
    \item Once the first block is pushed to a target, it pushes the second block to the remaining target.
\end{enumerate}
Thus combinatorially, the controller is capable of four different modes of behavior.
There are additional stochasticity in the controller behavior since there are many ways of pushing the same block into the same target.

The controller pushes the blocks to their targets following specific behavior primitives, such as moving to origin position, moving to a place collinear with a block and its target, and making a straight motion from that position towards the target unless the block rotates too much from its starting position.

Our models were trained on 1,000 demonstrations, all generated from the controller under the above randomized modes.

\paragraph{Franka kitchen environment:}

For the final set of experiments, we use the Franka Kitchen environment originally introduced in the Relay Policy Learning \cite{gupta2019relay} paper.
In that paper, the authors introduce a virtual kitchen environment where human participants in VR manipulated seven different objects in the kitchen: one kettle, one microwave, one sliding door, one hinged door, one light switch, and two burners.
In total, we use 566 demonstrations collected by the researchers in that paper, where in each demonstration episode, each participant performed four manipulation task specified by the researchers in advanced.

The manipulator agent in simulator is a Franka Emika Panda robot, which is controlled through a 9-dimensional action space controlling the robot's joint and end-effector position. The 60-dimensional observation space is split into two parts, the first 30 dimension contains information about the current position of the interesting factors in the environment, while the last 30 dimensions contain information about the goal of the demonstrator or the agent.
Note that in our demonstrations and our environments, we zero out the last 30 dimensions in all cases since we assume goal is not labelled in the demonstrations and is not specified in the unconditioned rollouts of the model.

One thing to note that, while the D4RL \cite{fu2020d4rl} paper also has three versions of the dataset, we chose to use the original version of the collected data from the Relay Policy Learning \cite{gupta2019relay} paper.
That is because the relay policy learning dataset is not labeled with intended tasks of the participants or rewards, while the D4RL dataset is geared towards that.

%% file: appendix_sections/bet_baselines.tex
\label{sec:app:baselines}
\paragraph{Multi-layer Perceptron with MSE} 
For our MLP with MSE baselines, we trained fully connected neural networks with optionally BatchNorm layers. In each of our environment, we varied the depth and the width of the MLPs to fit them best according to the bias-variance trade-off, while training them on 95\% of the dataset and testing on the remaining 5\% on the dataset in terms of MSE loss.

\paragraph{Nearest Neighbor}
Nearest Neighbor is conceptually the simplest baseline we show in this paper.
During training, our Nearest Neighbor model simply stores all the $(o, a)$ pairs.
During test time, given a query observation, $o$, we find the observation $o'$ with the minimum Euclidean distance to that in the representation space, and execute the associated action $a'$ in the environment.

While it is a simple baseline, we show that it has a surprisingly high effectiveness in simple environments like CARLA, or dense environments like Kitchen where there is less of a chance in going OOD simply by executing seen actions. On the other hand, in environments like Block-push where the model needs to interpolate or extrapolate more, the NN model fails more.

\paragraph{k-Nearest Neighbor with Locally Weighted Regression}
A slightly more robust version of NN for regression problems, k-NN with locally weighted regression or LWR, is the next baseline we use.
In this baseline, we take the k-nearest neighbors (in all our cases, $5$) in the observation representation space, and take a weighted average of their associated actions.
The weighting is based on the negative exponent of the distance, or namely, $\exp -||o - o'||$, as seen in \cite{pari2021surprising}.
This model is better than simple Nearest Neighbors in interpolations, and thus we see a higher success in the Kitchen environment.

\paragraph{Continuous Generative Model: VAE with Gaussian Prior} Following prior works\cite{spirl}, we use variational auto-encoders (VAE) for encoding and decoding sequences of actions into a smaller latent space.
The VAE here learns to compress a sequence of $T=10$ actions into a single latent variable $z$ of $10$ dimensions.
The hyperparameters for training the VAE has been taken directly from \citet{spirl}.

Concurrently with training the VAE, we train a state-conditioned latent prior model that tries to predict $P(z \mid o)$. 
This latent generator produces a vector of $\mu$ and $\sigma$ which is sampled to find latent $z$, and we feed a Gaussian distributed variable $z$ back into the decoder network where the action sequence is reconstructed.
For the current observation $o_t$, sequence of reconstructed actions $a_t, \cdots, a_{t+9}$ are performed in a simulated environment.

The design choices of this algorithm has been heavily inspired by \cite{spirl}. Although this model shows promise in theory, we found in practice that unconditional rollout from this model is not very successful.
We believe the shortcoming is a result of random sampling from the $z$ space that does not take into account the recently executed actions, and using a single-mode Gaussian as the state prior similar to \cite{spirl}, and thus this baseline is only slightly better than the MLP-MSE model.

\paragraph{Continuous Generative Models: Normalizing Flow with and without Prior}
Similar to \citet{singh2020parrot}, we use a Normalizing Flow \cite{dinh2016density} based generative model.
We follow the architectural choices and the hyperparameters from \cite{singh2020parrot} in our baseline implementation.

Our observation-conditioned Flow model is trained on the distribution $P(a \mid o)$ to continuously transform it into an identity Gaussian distribution of the same dimensions as $a$.
To find a better prior than simply an identity Gaussian, we also trained a prior model that generates $\mu, \sigma$ of a Gaussian distribution given the observation $o$.
We found that the prior improves the quality of the rollouts, however slightly.

We believe the under-performance of these continuous generative approaches were based on two major problems.
One is that they fail to take historical context in concern, and by being a continuous distribution, returned less likely actions that led to more rollouts going OOD.
Second, they were designed with a focus of making RL approachable by compressing the action space, which requires having a prior that is not so strict. However, most of \method{}'s performance comes from having a strong prior over the actions, which is only augmented by the action offset prediction.

\paragraph{Implicit Behavioral Cloning}
Implicit Behavioral Cloning (IBC) \cite{florence2021implicit} takes a different approach in behavioral cloning, where instead of learning a model $f(o) := a$, we learn an energy based model $E(o, a)$ where the intended action $a$ at any observation is defined as $\argmin_a E(o, a)$.
While this suffers from all the classic issues of training an EBM, like higher sample complexity and higher complexity in sampling, IBC models have been shown to have higher success in learning multi-modal and discontinuous actions.

As a baseline, we use the official implementation provided in  \url{https://github.com/google-research/ibc} For the CARLA environment, we use equivalent hyperparameters from the ``pushing from pixels'' hyperparameters. For the Block-pushing environment, we use the ``pushing from states'' hyperparameters. Finally, for the Kitchen environment, we use the ``D4RL kitchen'' hyperparameters.

While IBC is our strongest baseline, in our experience it is also one that is quite easy to overfit to our datasets. As a result, we monitored test performance over the training and had to employ early stopping for both the CARLA and the Block-pushing tasks.

\paragraph{Trajectory Transformers}
Trajectory Transformers~\cite{janner2021offline}, especially the variant that is trained without any rewards only on states and actions from demonstrations, seem similar to our approach, there are a few crucial differences. While we agree that BeT and Trajectory Transformer based behavior cloning both use some type of discretization to fit demonstration datasets with a minGPT, we believe that is where the similarities end. The primary differences between the algorithms is in our design choices: namely what distributions they model, and consequently how they treat the observations. The differences are explained more thoroughly below.

\begin{itemize}
    \item  \textbf{Modeled distribution}: From a provided set of demonstrations, trajectory transformers model the joint distribution P(action, observations). On the other hand, BeT models the conditional distribution P(action | observations). Modeling the joint distribution requires MinGPT to model the forward dynamics of the environment, which can be arbitrarily difficult based on the environment.
    \item \textbf{Observation discretization}: Because trajectory transformers have to model the observations as well, it needs to discretize the observation space. As a result, TT cannot extend to high dimensional observational spaces, such as visual observations. This limitation is also acknowledged by the authors of Trajectory Transformers. BeT, on the other hand, does not model the observations and thus does not need to discretize them. Thus BeT can scale to arbitrarily high dimensional observations, as we show in the CARLA environment experiments, where BeT learns behaviors from high dimensional visual observations.
    \item \textbf{Efficient historical encoding}: Trajectory transformer encodes each (state, action) pair into a total of $|S| + |A|$ input/output tokens, while BeT encodes them into one input/output token. On a base MinGPT implementation that means a $O( (|S| + |A|)^2 )$ efficiency gain for BeT, or for example 4761x less compute for the same historic context in the Kitchen environment. 
\end{itemize}

As a baseline, we trained and rolled out Trajectory Transformer on the Kitchen environment. It failed to complete any tasks for unconditioned, greedy, or beam search rollouts. 
We would like to note that the Kitchen environment is more complicated than the MuJoCo environments (HalfCheetah, Hopper, Walker2d, and Ant) that the paper experimented on.
At the same time, this environment has an order of magnitude fewer samples on the training set ($10^6$ vs. approximately $120$k).
We tried both our own implementation and the implementation from \url{https://github.com/Howuhh/faster-trajectory-transformer} with the recommended parameters for the AntMaze environment, which is the largest environment used by the authors.

%% file: appendix_sections/bet_compute_details.tex
\paragraph{Loss function details:}
In this paper, we use two loss functions that are inspired by practices in computer vision, in particular object detection.
The first of them is the Focal loss~\cite{lin2017focal}, and the second one is the Multi-task loss~\cite{girshick2015fast}.

The Focal loss is a simple modification over the cross entropy loss. While the normal cross entropy loss for binary classification can be thought of $\mathcal L_{ce} (p_t) = -\log (p_t)$, the Focal loss adds a term $(1-p_t)^\gamma$ to this, to make the new loss
\[\mathcal L_{focal} (p_t) = -(1-p_t)^\gamma\log (p_t)\]
This loss has the interesting property that its gradient is more steep for smaller values of $p_t$, while flatter for larger values of $p_t$. Thus, it penalizes and changes the model more for making errors in the low-probability classes, while is more lenient about making errors in the high probability classes. Using this error in the object detection world has helped with class imbalance between different classes, and here it helps \method{} learn to predict different $k$-means from the dataset even if their appearance in the dataset is not completely balanced.

For the multi-task loss, we use the formulation 
\[\text{MT-Loss}\left (\mathbf{a}, \left (\langle \hat a^{(j)}_i \rangle\right )_{j=1}^k\right ) = \sum_{j=1}^k \mathbb I [\lfloor \mathbf{a} \rfloor = j] \cdot \| \langle \mathbf{a} \rangle - \langle \hat a^{(j)} \rangle \|_2^2\]

This helps us penalize only the offset for the ground truth class, thus making sure the MinGPT is not trying to predict the right action offset through all classes and instead only trying to predict the action offset through the right class.

In practice, we optimize the combined loss, $\mathcal L_{focal} + \alpha \mathcal L_{mt}$ while $\alpha$ is a hyperparameter that just makes sure at initialization the two losses are of the same order of magnitude.
\paragraph{Compute details:}
All of our code was run in a single NVIDIA RTX 3080 GPU for state-based environments and RTX 8000 for image-based environments. 

\paragraph{Performance measurement details:}
We measured the performance reported in the Section~\ref{sec:ablations} in an NVIDIA RTX 3080 machine with AMD Threadripper 5950x CPUs. We took the average over three runs to minimize inter-run variances, and measured wall-clock time to report in the paper.

In terms of raw computation time to determine one action from the observations, in the Kitchen environment, BeT took 2.8 ms, while IBC took 52 ms and MLP, as the fastest point of comparison, took 0.5 ms. On the same environment, a single step of Trajectory Transformer took 867.86 ms, on an implementation that used more advanced tricks such as attention caching.

\paragraph{Hyperparameters list:}
We present the \method{} hyperparameters in Table~\ref{tab:bet-hparams} below:

\begin{table}[ht]
\caption{Environment-dependent hyperparameters in BeT.}
\label{tab:bet-hparams}
\centering
\begin{tabular}{lcccc}
Hyperparameter        & Point-mass & CARLA & Block-push & Kitchen \\
\hline
Layers                & 1          & 3     & 4          & 6       \\
Attention heads       & 2          & 4     & 4          & 6       \\
Embedding width       & 20         & 256   & 72         & 120     \\
Dropout probability   & 0.1        & 0.6   & 0.1        & 0.1     \\
Context size          & 2          & 10    & 5          & 10      \\
Training epochs       & 10         & 40    & 350        & 50      \\
Batch size            & 64         & 128   & 64         & 64      \\
Number of bins $k$    & 2;~3       & 32    & 24         & 64
\end{tabular}
\end{table}

However, we have found that as long as the model does not overfit, a wide range of parameters all yield favorable results for \method{}; thus, this table should be taken as reference values for reproducing our results rather than the only parameter sets that work.

Apart from that, we have some hyperparameters that are shared across all \method{} experiments. They are reproduced in Table \ref{tab:bet-shared-hparams}.

\begin{table}[ht]
\centering
\caption{Shared hyperparameters for BeT training}
\label{tab:bet-shared-hparams}
\begin{tabular}{ll}
Name               & Value       \\ \hline
Optimizer          & Adam        \\
Learning rate      & 1e-4        \\
Weight decay       & 0.1         \\
Betas              & (0.9, 0.95) \\
Gradient clip norm & 1.0         \\
\end{tabular}
\end{table}

%% file: appendix_sections/bet_pseudocode.tex
See the pseudocode described on Algorithm~\ref{alg:bet}.

\begin{algorithm}[tb]
   \caption{Learning Behavior Transformer from a dataset of behavior sequences.}
   \label{alg:bet}
\begin{algorithmic}
  \State \\ {\bfseries Input:} Dataset $(o_{t, i}, a_{t, i})_{t, i}$ for $ 0\le i \le$ number of demonstrations, $0 \le t \le$ maximum episode lengths, intended number of clusters $k$ and context history length $h$.
  \State \\ {\bfseries Initialize:} $\theta_M$ the parameters for MinGPT, $\{A_i\}_{i=1}^k$ cluster centers randomly in the action space.
  \State \\ {\bfseries Learn k-means encoder/decoder:}
  \State Using all possible $a_{t, i}$, learn the $k$ cluster centers using the $k$ means algorithm. 
  \State Set $\{A_i\}_{i=1}^k$ as the learned cluster centers.
  \State \\ {\bfseries Define functions:} 
  \State $\lfloor a \rfloor := \argmin_{i=1}^k ||a - A_i||$
  \State $\langle a \rangle := a - \lfloor a \rfloor$
  \State $\text{Enc}(a) := (\lfloor a \rfloor, \langle a \rangle)$
  \State $\text{Dec}(\lfloor a \rfloor, \langle a \rangle) := A_{\lfloor a \rfloor} + \langle a \rangle$

\State \\ {\bfseries Train MinGPT trunk of \method{}:}
\While {\textit{Not converged}} 
    \State Sample trajectory subsequence $(o_{t}, a_{t}), \cdots, (o_{t+h-1}, a_{t+h-1})$ from the dataset.
    \State Feed in the observations $(o_t, o_{t+1}, \cdots, o_{t+h-1})$ into the MinGPT.
    \State Get categorical distribution probabilities $p_{\tau,c}$ for $t \le \tau \le t+h-1, 1\le c \le k$.
    \State Compute focal loss $\mathcal L_{ce}$ of $p_{\tau,c}$ against ground truth class $\lfloor a_{\tau} \rfloor$, for all $\tau, c$ .
    \State Get the residual action offset per class, $\langle a_{\tau, c} \rangle$, for all $\tau, c$ from MinGPT.
    \State Calculate the multi-task loss, $\mathcal L_{mt}$, against true class predicted offset, $\sum_\tau || \langle a_{\tau, \lfloor a_\tau\rfloor} \rangle - \langle a_\tau \rangle||_2^2$
    \State Backprop using the normalized loss, $\mathcal L_{ce} + \alpha \mathcal L_{mt}$ where $\alpha$ makes the losses of equal magnitude.
\EndWhile
\State \\ {\bfseries Running on the environment:}
  \While {\textit{Episode not completed}} 
\State Stack the last $h$ observations in the environment, $(o_t, o_{t+1}, \cdots, o_{t+h-1})$ and feed into MinGPT.
\State Get categorical probabilities $p_{\tau,c}$ for $t \le \tau \le t+h-1, 1\le c \le k$ from the MinGPT.
\State Sample a class $c$ from $p_{t+h-1,c}$ for $1\le c \le k$.
\State Get the associated action offset, $\langle a_{t+h-1, c} \rangle$ from the MinGPT.
\State Decode into full continuous action, $\overline a_{t+h-1} := \text{Dec}(c, \langle a_{t+h-1, c} \rangle)$ 
\State Execute decoded action $\overline a_{t+h-1}$ into environment.
\EndWhile
\end{algorithmic}
\end{algorithm}

%% file: appendix_sections/bet_ablations.tex
In this section, we provide more details about the ablation studies presented in the main paper, as well as present detailed plots of our ablation studies that compare different versions of the \method{} architecture.

\subsection{Ablating historical context}
\label{sec:app:historical_context}
One of the reasons why we used transformer-based generative networks in our work is because of our hypothesis that having historical context helps our model learn better behavioral cloning.
Our experiments are performed by using the same model and simply providing sequences of length one on training and test time.
As we can see on Sec.~\ref{sec:ablations}, having some historical context helps our model learn much better.

\subsection{Ablating the number of discrete bin centers, $k$}
\label{sec:app:ablation:num_bins}
Since \method{} is trained with a sum of focal loss for the binning head and MSE loss for the offset head, the number of cluster centers present a trade-off in the architecture.
Concretely, as the number of bins go up, the log-likelihood loss goes up but the MSE loss goes down.
In Sec.~\ref{sec:ablations}, we showed that using only one bin ($k=1$) decreases the performance level of \method{}.

In this section, we present the plot of the variation in performance as $k$ value changes.

\begin{figure}[H]
    \centering
    \includegraphics[width=\linewidth]{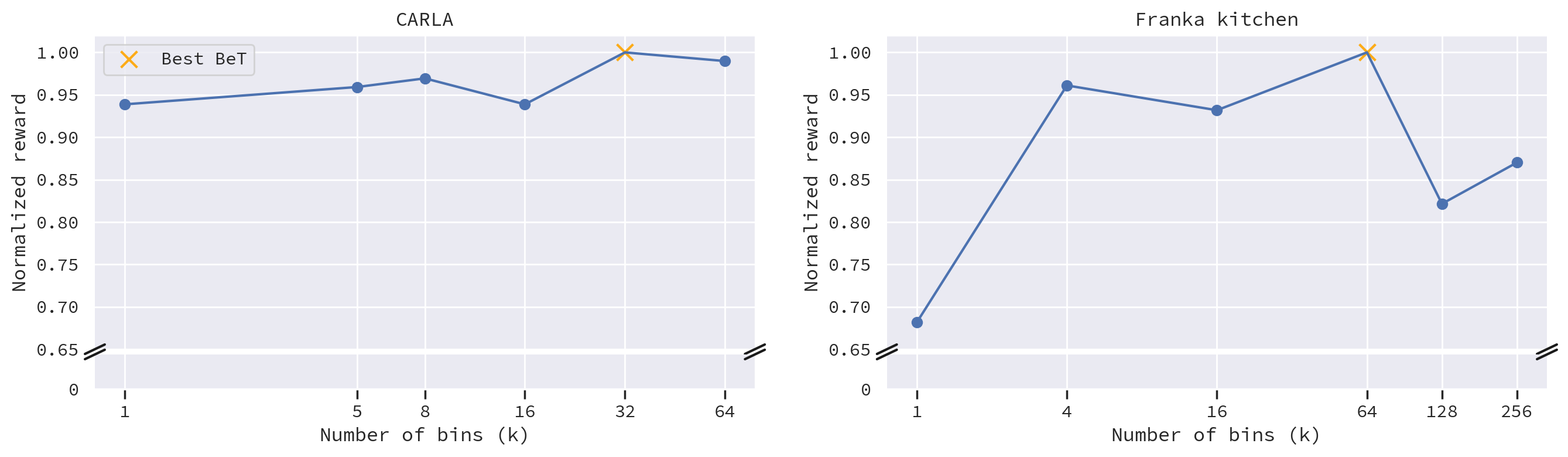}
    \caption{Ablating the number of discrete bin centers $k$ for \method{}. Reward is normalized with respect to the best performing model.}
    \label{fig:k_sweep}
    \vspace{-5pt}
\end{figure}

\subsection{Ablating the core model in the architecture}
\label{sec:app:ablation:transformer}

To ablate the core MinGPT transformer model in the architecture, we run three ablations, where we replace it respectively with a fully-connected multi-layer perceptron (MLP) network, a temporal convolution network, and an LSTM-based recurrent neural network.

\paragraph{Multi-Layer Perceptrons:}
Since generally MLP networks are not capable of taking in historical context in consideration, we instead stack the last $t$ frames of observation to pass into the MLP network.
Near the beginning of a trajectory, the stack of observation is zero-padded to $t$ frames.
For the intermediate layers in the MLP, we keep the same width and the number of layers as the corresponding MinGPT.

\paragraph{Temporal Convolution:}
Convolutions over the sequence length has been used in numerous prior works~\cite{oord2016wavenet,kalchbrenner2016neural,dauphin2017language,gehring2017convolutional,bai2018empirical} for sequence modeling.
As a baseline, we implement such temporal convolutional network to replace our MinGPT-based trunk.
We perform a temporal convolution over the same period of history that is provided to our transformer models.
We found that the performance of the temporal convolution models are constantly lower than our MinGPT based models.
However, temporal convolutional networks are easier to fit on our data compared to RNNs.

\paragraph{LSTM-based RNN:} Recurrent neural networks (RNNs) were the previous state-of-the-art for sequence modeling before transformer-based models.
In this work, we compare against an Long-short term memory (LSTM) \cite{gers1999learning} based RNN instead of a transformer based trunk.
We find that even with sufficient model capacity, the RNN based model took significantly longer than our MinGPT model to fit the same dataset.
Moreover, the quality of fit was worse, both in training and test time.
Finally, in open-ended rollouts, this performance downgrade is reflected in a far lower success rate for completing tasks in the environment (Table.~\ref{tab:ablation-table}).